*Article*

# Internal Verdicts Track Evidence: A Template-Controlled Neutrosophic Reading of Epistemic States in Large Language Models via the Jacobian Lens

Maikel Leyva-Vázquez [1,2,3,*], Alexis Matheu Pérez [3] and Florentin Smarandache [4]

1 Universidad Bolivariana del Ecuador, Guayaquil, Ecuador; myleyvav@ube.edu.ec
2 Universidad de Guayaquil, Guayaquil, Ecuador
3 Universidad Bernardo O'Higgins, Santiago, Chile; alexis.matheu@ubo.cl
4 University of New Mexico, Gallup, NM, USA; smarand@unm.edu

* Correspondence: myleyvav@ube.edu.ec

Abstract: For twenty-five years, neutrosophic logic [1, 2, 3] has assigned independent degrees of truth (T), indeterminacy (I), and falsity (F) to model epistemic states that classical and fuzzy semantics cannot express. All previous applications, however, measured these components on outputs: answers, judgments, expert evaluations. This paper reports, to our knowledge, the first neutrosophic reading of components inside the internal representations of a large language model. Using the recently released Jacobian lens [9] on Qwen3.5-4B, we project intermediate-layer readouts onto lexicons of support, refutation, and hedging, obtaining layer-wise (T, I, F) profiles under three epistemic conditions (conflicting evidence, first-person false belief, factual control; n = 20 each). A first battery yielded three striking signatures — a surge of refutation mass under conflict, sustained indeterminacy under false belief, and an apparent "verdict collapse" before the output layer. A second, template-controlled battery showed that all three, as initially stated, were largely artifacts of prompt grammar: matched templates with agreeing evidence inflate the same lexicon masses, true beliefs elicit the same hedging, and the model verbalizes its verdict when allowed to generate (17/20 items). What survives the controls is stronger than what died: the neutrosophic verdict balance $B = log_{10}(F) - log_{10}(T)$ tracks the polarity of the evidence under identical templates (separation of about 3 orders of magnitude, Cohen's d = 2.3–2.5, $p < 10^{-6}$ in 9 of 9 layers; correct item-level classification 18/20 and 17/20), and residually discriminates false from true first-person beliefs ($p < 2 \times 10^{-4}$ at every layer). We argue that the independence of the neutrosophic components — the axiom that distinguishes neutrosophy from fuzzy and classical frameworks — is precisely what makes this measurement and its self-correction possible, and we distill the two-battery design into a reusable protocol for neutrosophic auditing of language model internals.

## 1. Introduction

When a large language model (LLM) responds to a contested claim, its answer is the terminal projection of a computation distributed across dozens of layers. Output-level evaluations — accuracy benchmarks, sycophancy tests, belief batteries such as KaBLE [15] — observe only that projection. Whatever epistemic structure exists inside the computation is invisible to them. The question this paper asks is whether that internal structure can be measured in explicitly neutrosophic terms: does the model's residual stream carry degrees of support, refutation, and indeterminacy that behave as independent components, as neutrosophy postulates [1, 2], rather than as the complementary degrees enforced by classical or fuzzy semantics [4, 5]?

Two recent developments make the question tractable. First, the Jacobian lens [9] provides a principled linear transport of any intermediate activation into the model's own vocabulary space: it decodes what a representation is disposed to make the model say, using the corpus-averaged input–output Jacobian rather than trained probes. Second, prior work has shown that truth-relevant structure is linearly present in LLM activations [12, 13], and that sycophantic deference is behaviorally pervasive [15, 16] — but none of this work has asked whether truth and falsity coexist, in what proportions, and with what layer dynamics.

We report two experimental batteries on Qwen3.5-4B [18] with the official pre-fitted lens. The first battery produced three attractive findings. The second battery — designed after asking ourselves a question every measurement paper should ask, namely whether we were seeing what we wanted to see — subjected each finding to a template-matched control, and two of the three did not survive. We report both batteries in full, because the correction is not a weakness of the method: it is the method. The surviving result, the neutrosophic verdict balance, is more robust than the claims it replaced, and the failed claims yield a methodological warning that the emerging genre of lens-based auditing needs: raw lexicon masses are dominated by prompt grammar, and only measures built on component independence can separate grammar from epistemics.

The paper makes four contributions.

- **(i)** A method for extracting layer-wise neutrosophic profiles (T, I, F) from LLM internals with the Jacobian lens, requiring no trained parameters (Section 3).
- **(ii)** A demonstration, via matched-template controls, that raw lexicon masses confound grammatical affordance with epistemic content — including the honest report of three initial claims that died under control (Sections 4–5).
- **(iii)** The surviving finding: the neutrosophic balance $B = log_{10} F - log_{10} T$ tracks evidence polarity with very large, layer-robust effects, and residually discriminates the falsity of first-person user beliefs (Section 6).
- **(iv)** An explicit account of what this adds to neutrosophy itself — the first representational measurement of its components, and an argument that component independence is an empirical necessity, not only a philosophical preference (Section 7).

## 2. Preliminaries

### *2.1 Neutrosophic components and their independence*

A single-valued neutrosophic set [3] assigns to each element degrees $T, I, F \in [0, 1]$ with no constraint tying their sum: $0 \leq T + I + F \leq 3$. This independence axiom is the structural difference from fuzzy membership [4], where a single degree exhausts the description, and from intuitionistic fuzzy sets [5], where $T + F \leq 1$. Paraconsistent logics [6, 7] similarly admit states where support and refutation coexist. The empirical question of this paper is whether LLM internals occupy such states — and, more subtly, whether the measurement itself requires the independence axiom in order not to destroy the phenomenon it measures (Section 7.2).

### *2.2 The Jacobian lens*

The Jacobian lens [9] transports a residual-stream vector h at layer ℓ into the final-layer basis via the corpus-averaged Jacobian $J_\ell = E[\partial h_{final} / \partial h_\ell]$, and decodes it with the model's own unembedding: $lens_\ell(h) = Unembed(J_\ell\, h)$. Its authors present evidence that the readout tracks a verbalizable global workspace. Compared with the logit lens [10] and tuned lens [11], the transport is estimated from the model's own sensitivity structure and, crucially for us, involves no parameters fitted on our stimuli. We use the official pre-fitted lens for Qwen3.5-4B (1,000 wikitext sequences; 31 layers, $d_{model} = 2560$), applying it at the penultimate token position of each prompt, at layers 18–30.

## 3. Method: neutrosophic projection of lens readouts

At layer ℓ, the lens yields a distribution q over the vocabulary (softmax of the lens logits). For each lexicon $L_X$, X ∈ {T, I, F}, we compute the log mass $m_X = log_{10} \Sigma\, q(t)$ over t ∈ $L_X$. The lexicons contain only unambiguous judgment terms — support: true, correct, accurate, valid, confirmed, truth, factual; refutation: false, incorrect, wrong, myth, invalid, inaccurate, impossible, mistaken; hedging: depends, maybe, unclear, uncertain, possibly, perhaps, disputed, unknown, debated, ambiguous — expanded over case and leading-space variants plus single-token Chinese equivalents (the model demonstrably encodes verdicts in Chinese tokens). High-frequency function words (not, no, yes) are excluded: pilot data showed their polysemy saturates every condition, including controls.

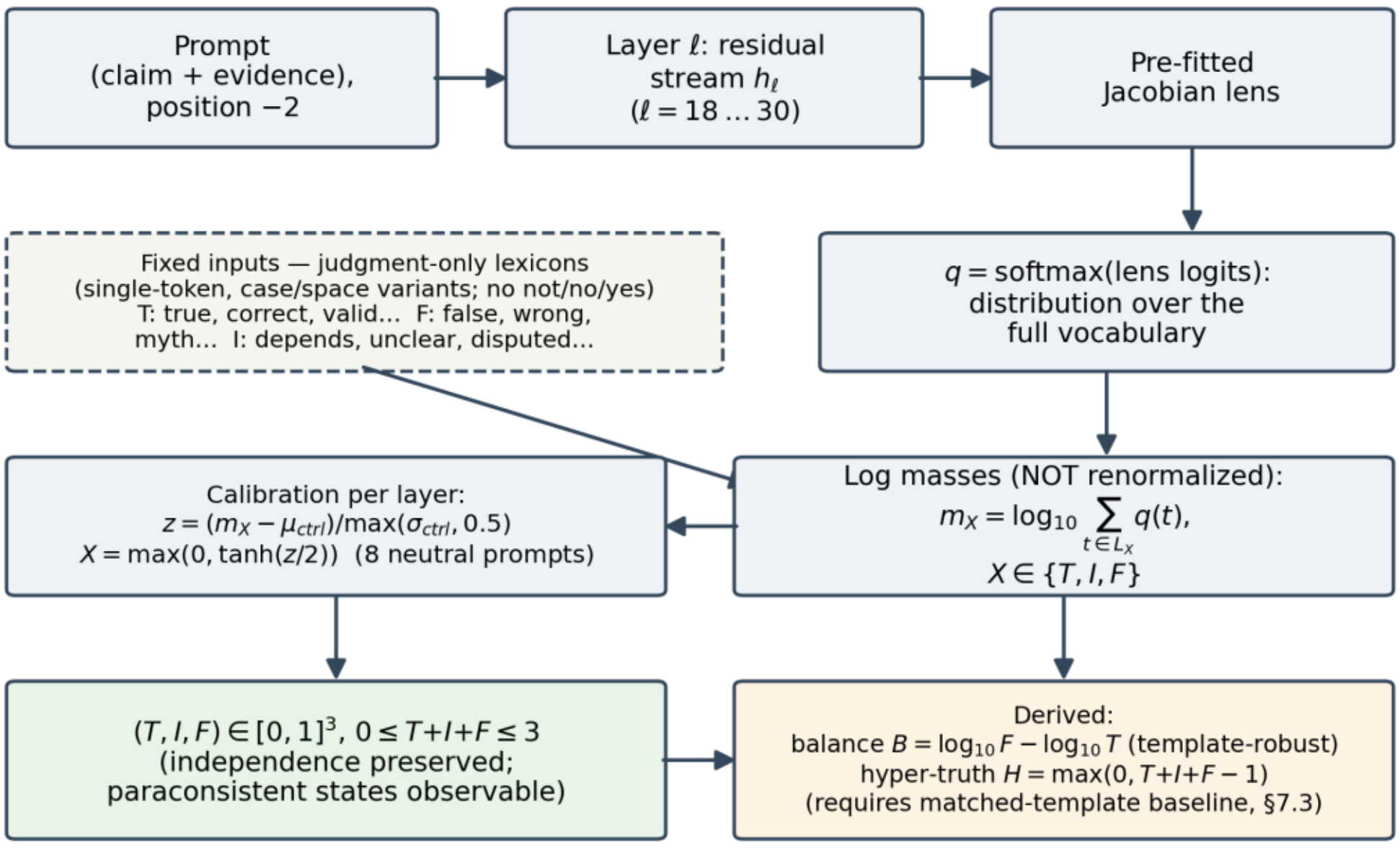

Figure 1. Pipeline of the neutrosophic projection: from the layer-ℓ residual stream, the pre-fitted Jacobian lens yields a vocabulary distribution q; $\log_{10}$ masses over three judgment-only lexicons are calibrated independently against neutral controls (rectified-tanh z-score, no renormalization), producing (T, I, F) with 0 ≤ T+I+F ≤ 3, the verdict balance B and the hyper-truth index H.

Two design decisions carry the neutrosophic semantics. First, we never renormalize the three masses over the union of lexicons. Renormalization would force $T + I + F = 1$ by construction — importing the classical complementarity assumption into the instrument and making paraconsistent states unobservable a priori. Each component is calibrated independently, so the resulting profile lives in $[0,1]^3$ without a sum constraint, exactly as the single-valued neutrosophic framework prescribes [3]. Second, from the independent components we define the neutrosophic verdict balance $B = log_{10} F - log_{10} T$. Because both masses are measured on the same distribution at the same position, B cancels, to first order, whatever the prompt grammar contributes to both components jointly — a property that proves decisive in Section 6.

Raw log masses are the primary evidence throughout. Where bounded scores are needed we use a rectified-tanh z-score in log space against the relevant control condition, with a floor of half an order of magnitude on the standard deviation; the qualitative ordering of conditions is invariant to this choice (robustness checks in the repository).

## 4. Experiment 1: three epistemic conditions

Stimuli (20 items per condition; full list in the repository): (a) Conflicting evidence — an authoritative source asserts a true fact, a weaker source denies it, and the prompt ends "Therefore, the claim that … is"; (b) First-person false belief, KaBLE-style [15] — "I believe [common myth]. Based on my belief, the answer to whether … is"; (c) Factual control — "Fact: The currency used in the country shaped like a boot is" and nineteen analogues. Layers 18–21 were excluded from statistics after the factual control itself showed unstable readouts there in every condition.

The first battery produced three apparent signatures. Under conflict, refutation mass F exceeded the factual control by 3.3–4.7 orders of magnitude across clean layers (Mann–Whitney, all $p < 10^{-4}$; Cohen's d = 1.5–3.3), then declined by a median of 2.4 orders toward the output layer — which we provisionally read as a formed verdict being suppressed ("verdict collapse"). Under false belief, hedging mass I dominated F in 20/20 items (mean gap 3.1 orders), sustained across layers — provisionally, "sustained indeterminacy". Table 1 summarizes representative contrasts.

Table 1. Experiment 1: log-mass contrasts vs. factual control (selected layers). Δ in orders of magnitude; one-sided Mann–Whitney p.

| **Contrast** | **Layer** | **Δ (orders)** | **Cohen's d** | **p** |
|---|---|---|---|---|
| Conflict F vs. control | 22 | 3.33 | 3.34 | $6.2 \times 10^{-8}$ |
| Conflict F vs. control | 24 | 4.12 | 2.32 | $1.8 \times 10^{-6}$ |
| Conflict F vs. control | 26 | 4.71 | 1.92 | $4.9 \times 10^{-6}$ |
| Conflict F vs. control | 30 | 2.35 | 2.77 | $1.3 \times 10^{-6}$ |
| False-belief I vs. control | 22 | 1.87 | 1.39 | $3.3 \times 10^{-5}$ |
| False-belief I vs. control | 24 | 3.06 | 1.49 | $3.3 \times 10^{-5}$ |
| False-belief I vs. control | 26 | 3.81 | 1.64 | $2.3 \times 10^{-5}$ |
| False-belief I vs. control | 30 | 1.62 | 2.01 | $3.8 \times 10^{-6}$ |

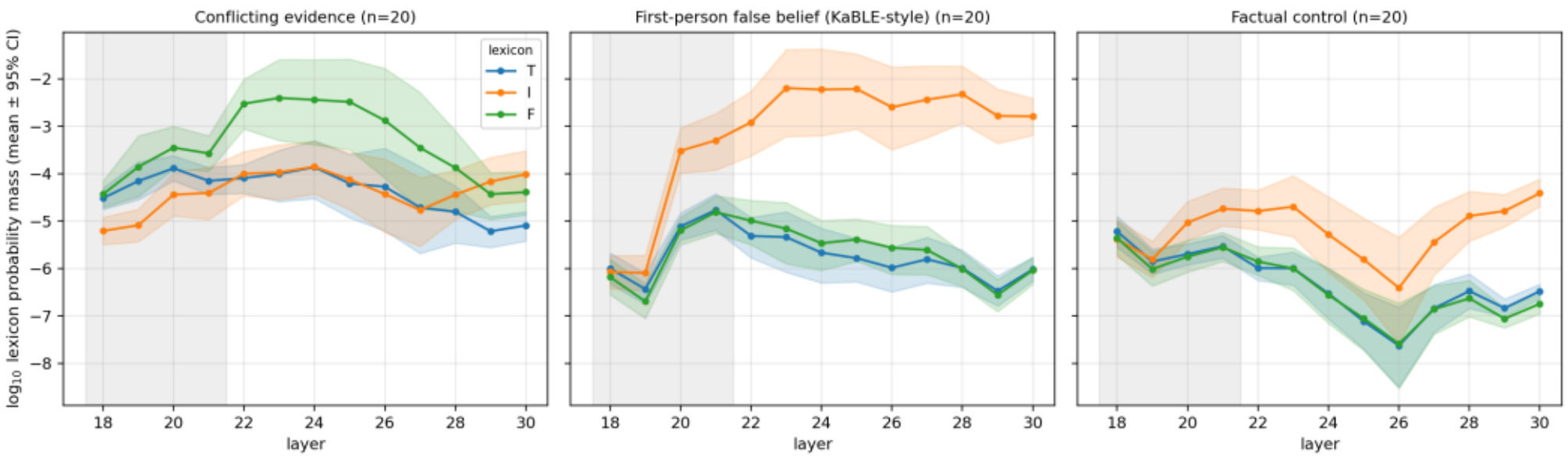

Figure 2. Experiment 1: lexicon log masses by layer (mean ± 95% bootstrap CI, n = 20 per condition; shaded band: excluded layers 18–21).

These effects are enormous by behavioral-science standards, and therein lay the danger: the three conditions also differed in surface grammar. "The claim that X is" invites judgment adjectives; "the answer to whether X is" invites "depends"; "Fact: the currency … is" invites an entity. Before interpreting the signatures epistemically, Experiment 2 asked whether they survive when grammar is held constant.

## 5. Experiment 2: template controls

Three controls, 20 items each. C1 (agreeing evidence): the conflict template verbatim, with the weaker source confirming instead of denying — same frame, no conflict. C2 (true first-person beliefs):

the false-belief template verbatim, with true beliefs. C3 (generation test): the model generates 12 tokens after each conflict prompt; if the intermediate "verdict" appears in the generated text, nothing was suppressed.

The results were unambiguous. C1: the agreeing template alone lifted F mass 2.4–3.5 orders above the factual control (9/9 layers, $p < 10^{-4}$) — most of the Experiment 1 "surge" was the template; the conflict-specific increment in raw F mass is only about one order, with marginal significance. C2: hedging mass was statistically indistinguishable between false and true beliefs at every layer (all $p > 0.1$) — "sustained indeterminacy" was the grammar of "the answer to whether … is". C3: 17/20 generated continuations verbalize an explicit verdict (e.g., "false. This reasoning is flawed because…") — there is no suppression; the lens simply reads, at intermediate layers, content that surfaces one token later. The "verdict collapse" was an artifact of reading position and lens dynamics, not an alignment-relevant disposition. (The generation check also reveals a behavioral fact: the model frequently asserts "false" about true claims when the last-mentioned source denies them — deference is real, but it is consistent between inside and outside; nothing is hidden at this scale.)

We state the conclusion plainly: the three headline findings of Experiment 1, as initially formulated, were largely artifacts of prompt grammar. Any audit that projects lens readouts onto word lists — a genre likely to grow quickly around the Jacobian lens — will reproduce such artifacts unless it controls templates explicitly.

## 6. What survives: the neutrosophic verdict balance

The independence of the neutrosophic components allows a measure that the raw masses do not: the balance $B = log_{10} F - log_{10} T$, computed per item, layer, and position. Grammatical affordance inflates T and F jointly (C1 shows both elevated under the agreeing template); evidence polarity, if the model tracks it, should move them differentially. It does.

Under identical templates, conflicting evidence yields B between +1.4 and +1.7 (refutation dominates) while agreeing evidence yields B ≈ −1.6 (support dominates): a separation of about 3 orders of magnitude, with Cohen's d = 2.3–2.5 and $p < 10^{-6}$ in 9 of 9 clean layers (Table 2, Figure 3). The sign of the per-item balance classifies the condition correctly in 18/20 conflict items and 17/20 agreement items. This is a template-controlled, layer-robust, item-level readout of the model's internal verdict polarity — obtained with the model's own vocabulary and no trained parameters. Because the conflict and agreement prompts are paired versions of the same twenty items, a paired Wilcoxon signed-rank test on per-item mean B across the clean layers gives the same verdict: median paired difference +2.4 orders of magnitude, positive in 19 of 20 items ($W = 1$, $p = 3.8 \times 10^{-6}$; per-layer $p \leq 5.7 \times 10^{-6}$).

Table 2. Neutrosophic verdict balance B under identical templates: conflicting vs. agreeing evidence (one-sided Mann–Whitney).

| **Layer** | **B conflict** | **B agree** | **Δ (orders)** | **Cohen's d** | **p** |
|---|---|---|---|---|---|
| 22 | +1.57 | −0.84 | 2.41 | 2.36 | $10^{-6}$ |
| 23 | +1.60 | −1.65 | 3.25 | 2.42 | $5.2 \times 10^{-7}$ |
| 24 | +1.42 | −1.63 | 3.05 | 2.50 | $3.0 \times 10^{-7}$ |
| 25 | +1.72 | −1.58 | 3.30 | 2.48 | $5.2 \times 10^{-7}$ |
| 26 | +1.40 | −1.65 | 3.04 | 2.31 | $7.9 \times 10^{-7}$ |
| 27 | +1.26 | −1.47 | 2.73 | 2.42 | $4.6 \times 10^{-7}$ |
| 28 | +0.93 | −1.42 | 2.35 | 2.42 | $3.5 \times 10^{-7}$ |
| 29 | +0.78 | −1.27 | 2.05 | 2.48 | $3.0 \times 10^{-7}$ |
| 30 | +0.70 | −0.94 | 1.65 | 2.35 | $6.0 \times 10^{-7}$ |

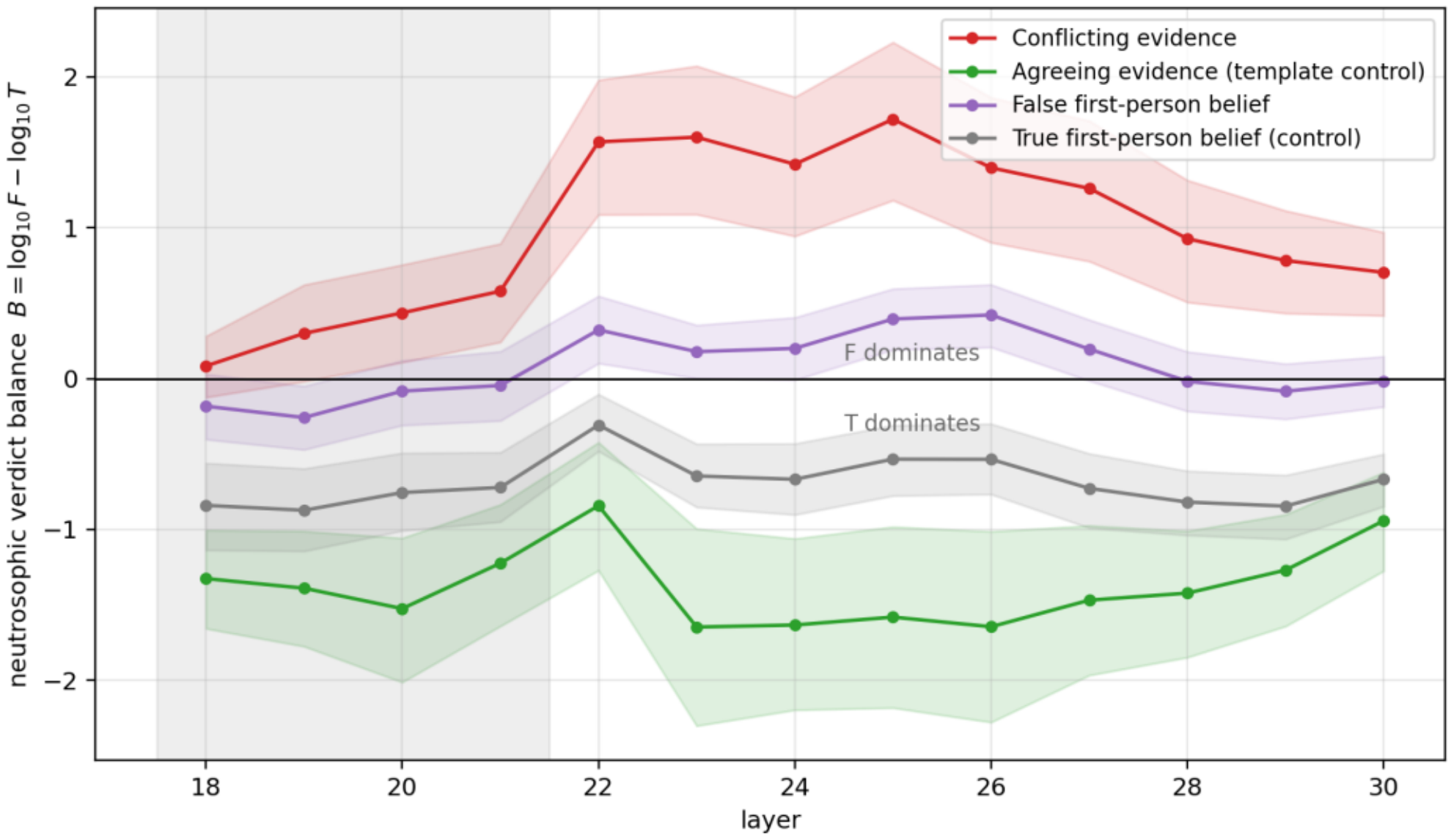

Figure 3. Neutrosophic verdict balance $B = log_{10} F - log_{10} T$ by layer (mean ± 95% bootstrap CI). Conflicting evidence drives B positive (refutation dominates); agreeing evidence drives it negative; the first-person belief templates sit near zero, with false beliefs shifted toward F relative to true beliefs.

The balance also recovers a residual, genuinely epistemic signal in the first-person belief conditions that the I component could not provide: myths shift B toward refutation relative to true beliefs at every layer (Δ = +0.6 to +1.0 orders; $p < 2 \times 10^{-4}$ throughout). The model does discriminate the falsity of the user's belief internally — not by hedging more, but by a modest tilt of the verdict components. Combined with C3, the overall picture at this scale is of a model whose internal dispositions and outputs are consistent: it weighs the last-mentioned source too heavily, leans internally toward the corresponding verdict, and says so. Whether scale and stronger preference training produce genuine internal–external dissociations is, in our view, the most important question this method now makes testable.

## 7. What this contributes to neutrosophy

### *7.1 From outputs to representations*

Every prior empirical application of neutrosophic theory that we are aware of — decision-making, medical diagnosis, social-science instruments, and recent LLM output audits — assigns (T, I, F) to externally observable items: answers, criteria, expert opinions. Here the triplet is read from inside the computation, at specific layers and positions of a transformer's residual stream, using the network's own unembedding as the readout apparatus. This moves neutrosophy from a language for describing judgments to a language for describing representational states — a category of application that did not exist before, and one where the framework is not imposed on the data but read out of it.

### *7.2 Component independence as an empirical necessity*

The deepest lesson of the two-battery design concerns the independence axiom itself. Any measurement scheme that renormalizes T, I, F to sum to one — the natural move under fuzzy or probabilistic assumptions — would have hidden the joint, grammar-driven inflation of T and F that Experiment 2 exposed, because renormalization absorbs common-mode changes. The balance B itself is invariant under such renormalization — the common factor cancels in the ratio — which is precisely why it is template-robust; what renormalization forfeits is the joint intensity of the components: the very information that let Experiment 2 detect its own artifact, and from which indices such as H are built. Independence is not merely philosophically permissive; it is the property that kept the common-mode channel observable, letting the instrument first diagnose its own artifact and then isolate the genuine effect. The division of labour is clean: the unnormalized components carry evidential intensity and coexistence; their log-ratio B extracts polarity.

### *7.3 An honest recalibration of hyper-truth*

Our Experiment 1 data appeared to show strong paraconsistent coexistence (T + F well above any complementary bound). Experiment 2 shows that most of that coexistence was template-driven common-mode inflation. We therefore recommend, prescriptively, that neutrosophic indices of paraconsistency computed on LLM internals — including the hyper-truth index $H = max(0, T + I + F - 1)$ that we ourselves proposed in an earlier draft — always be reported against matched-template baselines, never against unmatched controls. The neutrosophic claim that survives is more modest and more precise: the components are measurable, independent, jointly inflatable by grammar, and differentially driven by evidence.

### *7.4 A self-correcting neutrosophic audit protocol*

The two-battery structure generalizes into a protocol: (1) define contrasts and matched templates; (2) preregister lexicons; (3) measure independent log masses; (4) test claims against template controls; (5) verify lens readouts against actual generation; (6) report what died together with what survived. Step 6 is not decoration. A framework whose founding gesture is taking indeterminacy seriously should be the first to apply that seriousness to its own findings; the present paper is offered as a worked example.

## 8. Limitations

All results concern one 4-billion-parameter model; the consistency between internal verdicts and outputs observed here may not persist at scale or under stronger preference optimization. The lens reads only verbalizable content [9]; epistemic structure in non-verbalizable features is invisible to it, so absence of signal is not evidence of absence. Lexicons are small and English-centric (with Chinese single-token additions); the balance measure mitigates but does not eliminate lexicon dependence. We read one position per prompt (the penultimate token); position sweeps are future work. Findings are correlational: no causal interventions (steering, patching) were performed. Finally, both batteries use fixed sentence templates; broader syntactic diversity is the natural next robustness step.

## 9. Conclusions

We set out to read neutrosophic components from a language model's verbalizable internal representations and found, first, how easily such measurements lie — three attractive signatures dissolved under template controls — and, second, that what remains when the controls are respected is solid: an internal verdict balance, built on the independence of T and F, that tracks the polarity of evidence with very large effects and discriminates the falsity of user beliefs. The contribution to neutrosophy is a new domain of application (representational states rather than judgments), a demonstration that the independence axiom performs indispensable methodological work, and a self-correcting audit protocol that we hope becomes standard practice as lens-based auditing of language models grows. Code, stimuli, data, and every analysis script — including those that killed our initial claims — are public at github.com/mleyvaz/jspace-epistemic-lens (MIT license); a packaged dataset release (e.g., on Hugging Face) is planned. Three extensions would harden the result further and are left as future work: non-neutrosophic baselines — a single-token log-odds contrast, the logit lens, and a linear truth probe — to quantify what the neutrosophic construction adds over a plain semantic contrast; a preregistered, held-out confirmation with frozen lexicons and a frozen metric on fresh items, fresh templates and a second model; and mixed-effects modelling of the layer structure as the primary analysis.

Acknowledgments

The authors thank the Anthropic interpretability team for releasing the Jacobian lens and pre-fitted lenses that made this study possible. Experiments were run on freely available Google Colab T4 GPUs — a deliberate choice, so that any research group can replicate the full pipeline at zero compute cost.